\documentclass[10pt,twocolumn,letterpaper]{article}

\usepackage{amssymb}  
\usepackage{graphicx}
\usepackage{color}
\usepackage{rotating}
\usepackage{cvpr}
\usepackage{times}
\usepackage{epsfig}
\usepackage{amsmath}
\usepackage{arydshln}
\usepackage{nopageno}

\usepackage{authblk} 

\usepackage[breaklinks=true,bookmarks=false]{hyperref}

\cvprfinalcopy 

\def\cvprPaperID{3473} 

\begin{document}

\title{Deep ChArUco: Dark ChArUco Marker Pose Estimation}

\author{Danying Hu}
\author{Daniel DeTone}
\author{Vikram Chauhan}
\author{Igor Spivak}
\author{Tomasz Malisiewicz}
\affil{Magic Leap, Inc. \protect\\
\texttt{\{dhu,ddetone,vchauhan,ispivak,tmalisiewicz\}@magicleap.com}}

\maketitle
\begin{abstract}
\vspace{-.1in}
ChArUco boards are used for camera calibration, monocular pose estimation, and pose verification in both robotics and augmented reality. Such fiducials are detectable via traditional computer vision methods (as found in OpenCV) in well-lit environments, but classical methods fail when the lighting is poor or when the image undergoes extreme motion blur. We present Deep ChArUco, a real-time pose estimation system which combines two custom deep networks, ChArUcoNet and RefineNet, with the Perspective-n-Point (PnP) algorithm to estimate the marker's 6DoF pose. ChArUcoNet is a two-headed marker-specific convolutional neural network (CNN) which jointly outputs ID-specific classifiers and 2D point locations. The 2D point locations are further refined into subpixel coordinates using RefineNet. Our networks are trained using a combination of auto-labeled videos of the target marker, synthetic subpixel corner data, and extreme data augmentation. We evaluate Deep ChArUco in challenging low-light, high-motion, high-blur scenarios and demonstrate that our approach is superior to a traditional OpenCV-based method for ChArUco marker detection and pose estimation.
\end{abstract}

\vspace{-.16in}
\section{Introduction}

\begin{figure}
  \includegraphics[width=\linewidth]{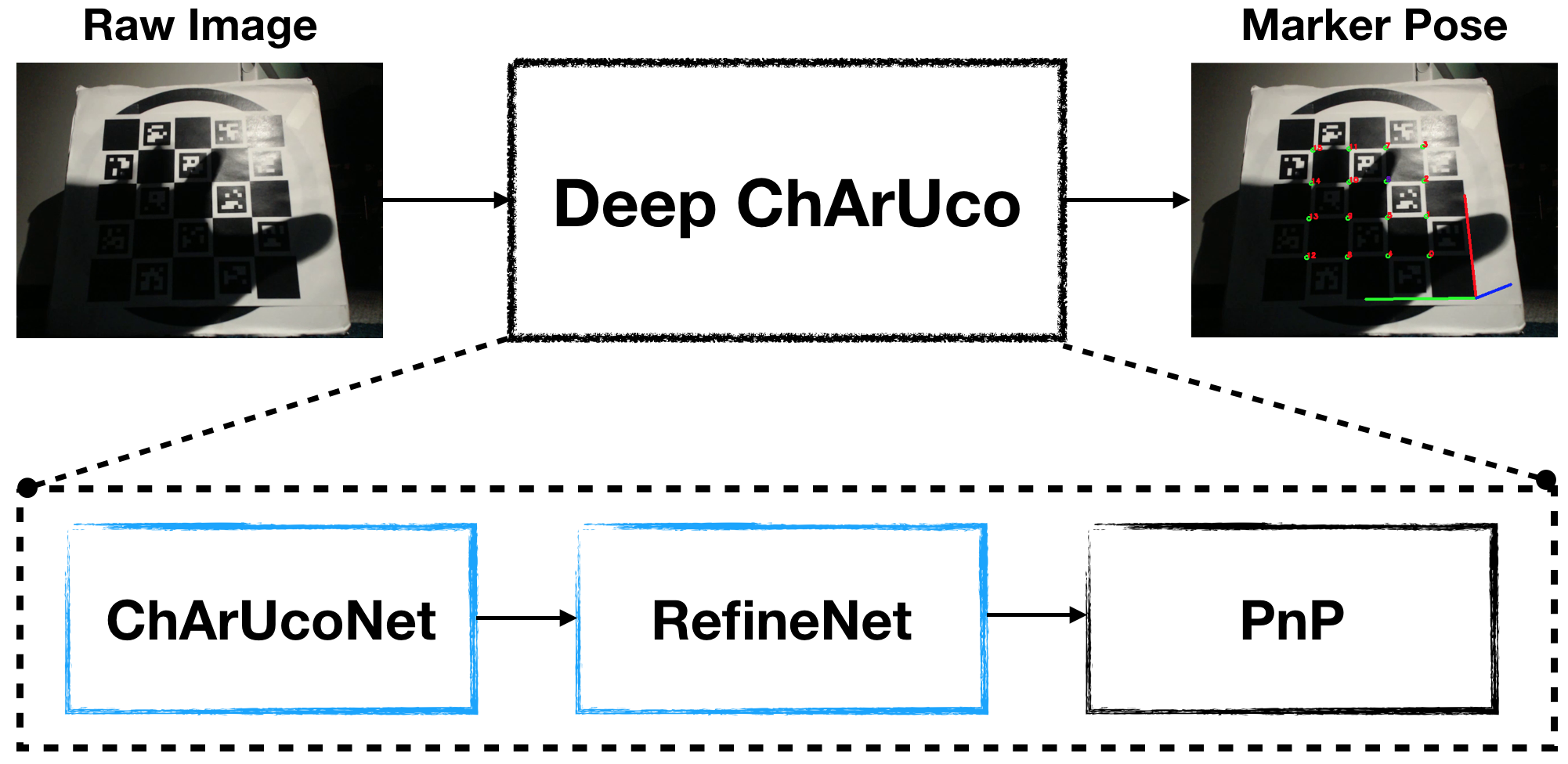}
  \caption{{\bf Deep ChArUco is an end-to-end system for ChArUco marker pose estimation from a single image}. Deep ChArUco is composed of ChArUcoNet for point detection (Section~\ref{sec:charuconet}), RefineNet for subpixel refinement (Section~\ref{sec:refinenet}), and the Perspective-n-Point (PnP) algorithm for pose estimation (Section~\ref{sec:pnp}). \emph{For this difficult image, OpenCV does not detect enough points to determine a marker pose.}}
  \label{fig:teaser}
  \vspace{-.2in}
\end{figure}

In this paper, we refer to computer-vision-friendly 2D patterns that are unique and have enough points for 6DoF pose estimation as \emph{fiducials} or \emph{markers}. ArUco markers~\cite{aruco,garrido2014} and their derivatives, namely ChArUco markers, are frequently used in augmented reality and robotics. For example, Fiducial-based SLAM~\cite{fiducialsfm,fiducialslam2} reconstructs the world by first placing a small number of fixed and unique patterns in the world. The pose of a calibrated camera can be estimated once at least one such marker is detected. But as we will see, traditional ChArUco marker detection systems are surprisingly frail. In the following pages, we motivate and explain our recipe for creating a state-of-the-art Deep ChArUco marker detector based on deep neural networks.

We focus on one of the most popular class of fiducials in augmented reality, namely ChArUco markers. In this paper, we highlight the scenarios under which traditional computer vision techniques fail to detect such fiducials, and present \emph{Deep ChArUco}, a deep convolutional neural network system trained to be accurate and robust for ChArUco marker detection and pose estimation (see Figure~\ref{fig:teaser}). The main {\bf contributions} of this work are:
\vspace{-.05in}
\begin{enumerate}  
\item A state-of-the-art and real-time marker detector that improves the robustness and accuracy of ChArUco pattern detection under extreme lighting and motion
\vspace{-.05in}
\item Two novel neural network architectures for point ID classification and subpixel refinement
\vspace{-.05in}
\item A novel training dataset collection recipe involving auto-labeling images and synthetic data generation 
\end{enumerate}
\vspace{-.05in}

{\bf Overview:} We discuss both traditional and deep learning-based related work in Section~\ref{sec:related}. We present ChArUcoNet, our two-headed custom point detection network, and RefineNet, our corner refinement network in Section~\ref{sec:network}. Finally, we describe both training and testing ChArUco datasets in Section~\ref{sec:dataset}, evaluation results in Section~\ref{sec:eval}, and conclude with a discussion in Section~\ref{sec:conclusions}. 

\vspace{-.1in}
\section{Related Work}
\label{sec:related}

\subsection{Traditional ChArUco Marker Detection}
\begin{figure}[ht]
  \centering
  \includegraphics[width=.64\linewidth]{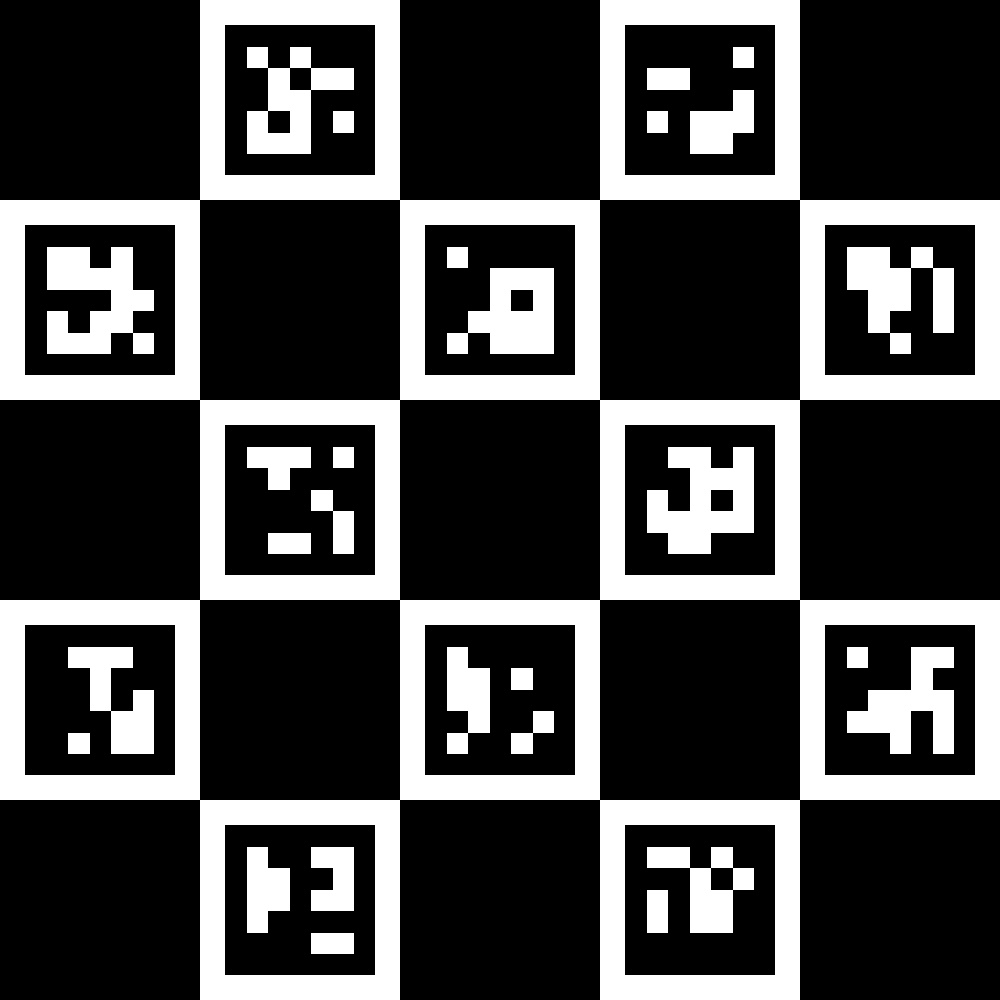}
  \caption{{\bf ChArUco = Chessboard + ArUco}. Pictured is a 5x5 ChArUco board which contains 12 unique ArUco patterns. For this exact configuration, each 4x4 chessboard inner corner is assigned a unique ID, ranging from $0$ to $15$. The goal of our algorithm is to detect these unique $16$ corners and IDs.}
  \vspace{-.1in}
  \label{fig:charuco}
\end{figure}

A ChArUco board is a chessboard with ArUco markers embedded inside the white squares (see Figure~\ref{fig:charuco}). ArUco markers are modern variants of earlier tags like ARTag~\cite{artag} and AprilTag~\cite{apriltag}. A traditional ChArUco detector will first detect the individual ArUco markers. The detected ArUco markers are used to interpolate and refine the position of the chessboard corners based on the predefined board layout. Because a ChArUco board will generally have $10$ or more points, ChArUco detectors allow occlusions or partial views when used for pose estimation. In the classical OpenCV method~\cite{bradski2000opencv}, the detection of a given ChArUco board is equivalent to detecting each chessboard inner corner associated with a unique identifier. In our experiments, we use the $5 \times 5$ ChArUco board which contains the first $12$ elements of the \verb+DICT_5x5_50+ ArUco dictionary as shown in Figure~\ref{fig:charuco}.

\subsection{Deep Nets for Object Detection}
Deep Convolutional Neural Networks have become the standard tool of choice for object detection since 2015 (see systems like YOLO~\cite{yolo}, SSD~\cite{Wei2016}, and Faster R-CNN~\cite{Ren2015}). While these systems obtain impressive multi-category object detection results, the resulting bounding boxes are typically not suitable for pose inference, especially the kind of high-quality 6DoF pose estimation that is necessary for augmented reality. More recently, object detection frameworks like Mask-RCNN~\cite{maskrcnn} and PoseCNN~\cite{posecnn} are building pose estimation capabilities directly into their detectors.

\subsection{Deep Nets for Keypoint Estimation}
Keypoint-based neural networks are usually fully-convolutional and return a set of skeleton-like points of the detected objects. Deep Nets for keypoint estimation are popular in the human pose estimation literature. Since for a rigid object, as long as we can repeatably detect a smaller yet sufficient number of 3D points in the 2D image, we can perform PnP to recover the camera pose. Albeit indirectly, keypoint-based methods do allow us to recover pose using a hybrid deep (for point detection) and classical (for pose estimation) system. One major limitation of most keypoint estimation deep networks is that they are too slow because of the expensive upsampling operations in hourglass networks~\cite{hourglass}. Another relevant class of techniques is those designed for human keypoint detection such as faces, body skeletons~\cite{cao2017realtime}, and hands~\cite{simon2017hand}.

\begin{figure}[h]
\centering
\includegraphics[width=.5\textwidth]{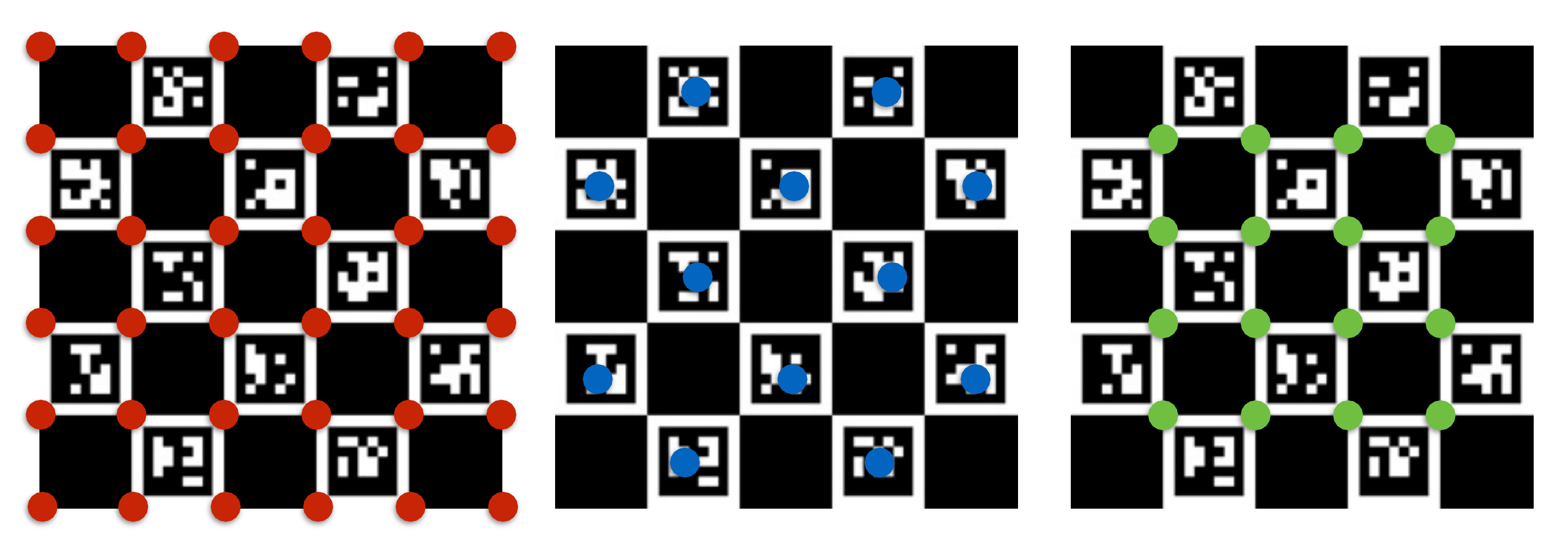}
\caption{{\bf Defining ChArUco Point IDs}. These three examples show different potential structures in the pattern that could be used to define a single ChArUco board. a) Every possible corner has an ID. b) Interiors of ArUco patterns chosen as IDs. c) Interior chessboard of 16 ids, from id 0 of the bottom left corner to id 15 of the top right corner ({\bf our solution}).}
\label{fig:defining-corners}
\vspace{-.1in}
\end{figure}

\begin{figure*}[t!]
\vspace{-.2in}
\begin{center}
\includegraphics[width=\linewidth]{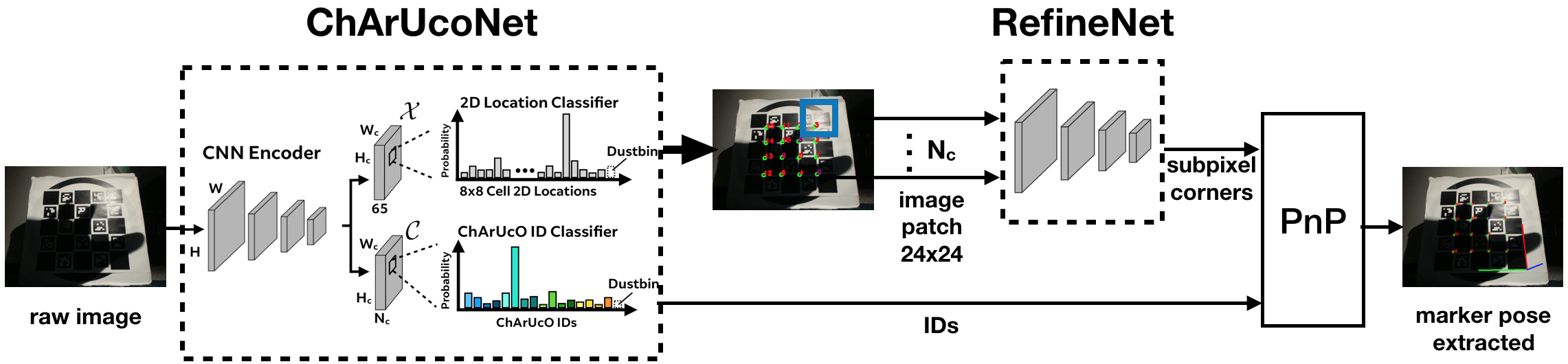}
\end{center}
\caption{{\bf Two-Headed ChArUcoNet and RefineNet.} ChArUcoNet is a SuperPoint-like~\cite{detone2018} network for detecting a specific ChArUco board. Instead of a descriptor head, we use a point ID classifier head. One of the network heads detects 2D locations of ChArUco boards in $\mathcal{X}$ and the second head classifies them in $\mathcal{C}$. Both heads output per-cell distributions, where each cell is an 8x8 region of pixels. We use $16$ unique points IDs for our 5x5 ChArUco board. ChArUcoNet's output is further refined via a RefineNet to obtain subpixel locations. 
\label{fig:charuconet-heads} } 
\end{figure*}

\subsection{Deep Nets for Feature Point Detection}
The last class of deep learning-based techniques relevant to our discussion is deep feature point detection systems--methods that are deep replacements for classical systems like SIFT~\cite{sift} and ORB~\cite{orb}. Deep Convolutional Neural Networks like DeTone et al's SuperPoint system~\cite{detone2018} are used for joint feature point and descriptor computation. SuperPoint is a single real-time unified CNN which performs the roles of multiple deep modules inside earlier deep learning for interest-point systems like the Learned Invariant Feature Transform (LIFT)~\cite{Moo2016}. Since SuperPoint networks are designed for real-time applications, they are a starting point for our own Deep ChArUco detector. 

\section{Deep ChArUco: A System for ChArUco Detection and Pose Estimation}
\label{sec:network}
In this section, we describe the fully convolutional neural network we used for ChArUco marker detection. Our network is an extension of SuperPoint~\cite{detone2018} which includes a custom head specific to ChArUco marker point identification. We develop a multi-headed SuperPoint variant, suitable for ChArUco marker detection (see architecture in Figure~\ref{fig:charuconet-heads}). Instead of using a descriptor head, as was done in the SuperPoint paper, we use an id-head, which directly regresses to corner-specific point IDs. We use the same point localization head as SuperPoint -- this head will output a distribution over pixel location for each 8x8 pixel region in the original image. This allows us to detect point locations at full image resolution without using an explicit decoder.

{\bf Defining IDs.} In order to adapt SuperPoint to ChArUco marker detection, we must ask ourselves: \emph{which points do we want to detect?}
In general, there are multiple strategies for defining point IDs (see Figure~\ref{fig:defining-corners}). For simplicity, we decided to use the 4x4 grid of interior chessboard corners for point localization, giving a total of $16$ different point IDs to be detected. The ID classification head will output distribution over $17$ possibilities: a cell can belong to one of the $16$ corner IDs or an additional ``dustbin'' none-of-the-above class. This allows a direct comparison with the OpenCV method since both classical and deep techniques attempt to localize the same $16$ ChArUco board-specific points.

\subsection{ChArUcoNet Network Architecture}
\label{sec:charuconet}
The ChArUcoNet architecture is identical to that of the SuperPoint~\cite{detone2018} architecture, with one exception - the descriptor head in the SuperPoint network is replaced with a ChArUco ID classification head $\mathcal{C}$ as shown in Figure~\ref{fig:charuconet-heads}.

The network uses a VGG-style encoder to reduce the dimensionality of the image. The encoder consists of 3x3 convolutional layers, spatial downsampling via pooling and non-linear activation functions. There are three max-pooling layers which each reduce the spatial dimensionality of the input by a factor of two, resulting in a total spatial reduction by a factor of eight.  The shared encoder outputs features with spatial dimension $H_c \times W_c$. We define $H_c = H / 8$ and $W_c = W / 8$ for an image sized $H \times W$. The keypoint detector head outputs a tensor $\mathcal{X}\in \mathbb{R}^{H_c\times W_c \times 65}$. Let $N_{c}$ be the number of ChArUco points to be detected (e.g. for a 4x4 ChArUco grid $N_c = 16$). The ChArUco ID classification head outputs a classification tensor $\mathcal{C} \in \mathbb{R}^{H_c\times W_c \times (N_c+1)}$ over the $N_c$ classes and a dustbin class, resulting in $N_c + 1$ total classes. The ChArUcoNet network was designed for speed--the network weights take $4.8$ Megabytes and the network is able to process $320 \times 240$ sized images at approximately 100fps  using an NVIDIA\textsuperscript{\textregistered} GeForce GTX 1080 GPU. 

\subsection{RefineNet Network Architecture}
\label{sec:refinenet}
To improve pose estimation quality, we additionally perform \emph{subpixel localization} -- we refine the detected integer corner locations into subpixel corner locations using RefineNet, a deep network trained to produce subpixel coordinates. RefineNet, our deep counterpart to OpenCV's \verb+cornerSubPix+, takes as input a $24 \times 24$ image patch and outputs a single subpixel corner location at $8 \times$ the resolution of the central $8\times 8$ region. RefineNet performs softmax classification over an $8 \times$ enlarged central region -- RefineNet finds the peak inside the $64 \times 64$ subpixel region (a $4096$-way classification problem). RefineNet weights take up only $4.1$ Megabytes due to a bottleneck layer which converts the $128$D activations into $8$D before the final $4096$D mapping. Both ChArUcoNet and RefineNet use the same VGG-based backbone as SuperPoint~\cite{detone2018}.

For a single imaged ChArUco pattern, there will be at most $16$ corners to be detected, so using RefineNet is as expensive as 16 additional forward passes on a network with $24 \times 24$ inputs.

\subsection{Pose Estimation via PnP}
\label{sec:pnp}

Given a set of 2D point locations and a known physical marker size we use the Perspective-n-Point (PnP) algorithm~\cite{hz} to compute the ChArUco pose w.r.t the camera. PnP requires knowledge of $K$, the camera intrinsics, so we calibrate the camera before collecting data. We calibrated the camera until the reprojection error fell below $0.15$ pixels. We use OpenCV's \verb+solvePnPRansac+ to estimate the final pose in our method as well as in the OpenCV baseline. 


\vspace{-.05in}
\section{ChArUco Datasets}
\begin{figure}[ht]
\vspace{-.1in}
\centering
\begin{tabular}{p{0.05cm}p{8.3cm}}
\rotatebox[origin=l]{90}{no data aug} &
{\mbox{\includegraphics[width=0.3\linewidth]{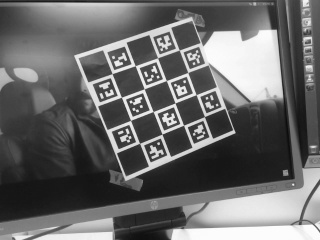}}
\mbox{\includegraphics[width=0.3\linewidth]{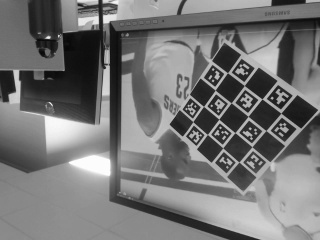}}
\mbox{\includegraphics[width=0.3\linewidth]{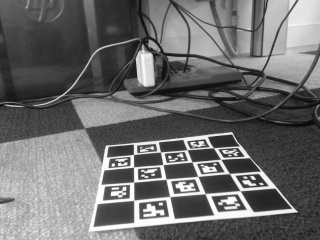}}
}\\
\rotatebox[origin=l]{90}{+data aug} &
{\mbox{\includegraphics[width=0.3\linewidth]{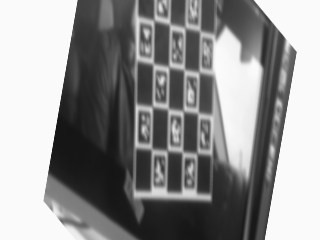}}
\mbox{\includegraphics[width=0.3\linewidth]{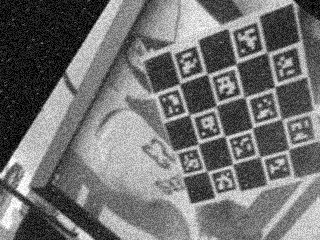}}
\mbox{\includegraphics[width=0.3\linewidth]{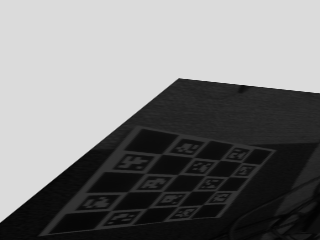}}
}
\end{tabular}
\caption{{\bf ChArUco Training Set.} Examples of ChArUco dataset training examples, before and after data augmentation.}
\label{fig:training-set}
\end{figure}

\vspace{-.05in}
\label{sec:dataset}
To train and evaluate our Deep ChArUco Detection system, we created two ChArUco datasets. The first dataset focuses on diversity and is used for training the ChArUco detector (see Figure~\ref{fig:training-set}). The second dataset contains short video sequences which are designed to evaluate system performance as a function of illumination (see Figure~\ref{fig:evaluation-set}).

\vspace{-.05in}
\subsection{Training Data for ChArUcoNet}
\vspace{-.05in}
We collected $22$ short video sequences from a camera with the ChArUco pattern in a random but static pose in each video. Some of the videos include a ChArUco board taped to a monitor with the background changing, and other sequences involve lighting changes (starting with good lighting). Videos frames are extracted into the positive dataset with the resolution of $320 \times 240$, resulting in a total of $7,955$ gray-scale frames. Each video sequence starts with at least 30 frames of good lighting. The ground truth of each video is auto-labeled from the average of the first 30 frames using the classical OpenCV method, as the OpenCV detector works well with no motion and good lighting.

The negative dataset contains $91,406$ images in total, including $82,783$ generic images from the MS-COCO dataset~\footnote{MS-COCO 2014 train: http://images.cocodataset.org/zips/train2014.zip} and $8,623$ video frames collected in the office. Our in-office data contains images of vanilla chessboards, and adding them to our negatives was important for improving overall model robustness. 

We collect frames from videos depicting ``other'' ChArUco markers (i.e., different than the target marker depicted in Figure~\ref{fig:charuco}). For these videos, we treated the classifier IDs as negatives but treated the corner locations as ``ignore.''

\begin{figure}[ht]
\vspace{-.2in}
\centering
\includegraphics[width=.48\textwidth]{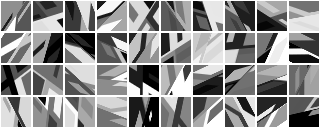}
\caption{{\bf RefineNet Training Images}. $40$ examples of synthetically generated image patches for training RefineNet.}
\vspace{-.1in}
\label{fig:synthetic-patches}
\end{figure}

\vspace{-.05in}
\subsection{Data Augmentation for ChArUcoNet}
\vspace{-.05in}
With data augmentation, each frame will undergo a random homographic transform and a set of random combination of synthetic distortions under certain probability (see Table~\ref{tab:distort}) during the training stage, which dramatically increases the diversity of the input dataset. The order and the extent of the applied distortion effects are also randomly selected for each frame. For example, Figure~\ref{fig:training-set} shows frames from the training sequences (top row) and augmented with a set of distortions (bottom row).

\begin{table}[h!]
\centering
\scalebox{0.85}{
\begin{tabular}{l|c}
\hline 
\textbf{Effect} &  \textbf{Probability} \\
\hline\hline
additive Gaussian noise & 0.5 \\
motion blur & 0.5 \\
Gaussian blur & 0.25 \\
speckle noise & 0.5 \\
brightness rescale & 0.5 \\
shadow or spotlight effect & 0.5 \\
homographic transform & 1.0 (positive set) / 0.0 (negative set) \\
\hline
\end{tabular}}
\vspace{.1in}
\caption{{\bf Synthetic Effects Applied For Data Augmentation}. During training we transform the images to capture more illumination and pose variations.}
\label{tab:distort}
\end{table}

\vspace{-.1in}
\subsection{Synthetic Subpixel Corners for RefineNet}
We train RefineNet using a large database of synthetically generated corner images. Each synthetic training image is $24 \times 24$ pixels and contains exactly one a ground-truth corner within the central $8 \times 8$ pixel region. For examples of such training image patches, see Figure~\ref{fig:synthetic-patches}.
 
\subsection{Evaluation Data} \label{sec:eval_data}
\begin{figure}[b]
\centering
{\mbox{\includegraphics[width=0.32\linewidth]{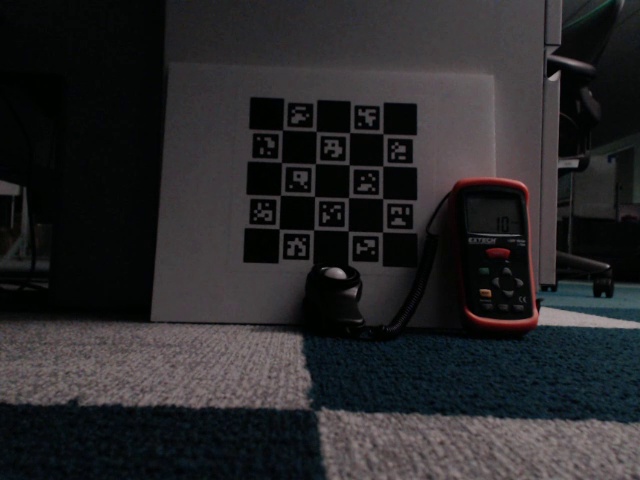}}
\mbox{\includegraphics[width=0.32\linewidth]{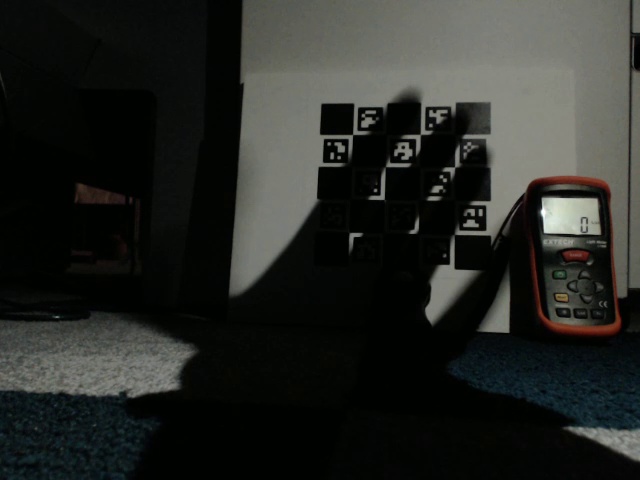}}
\mbox{\includegraphics[width=0.32\linewidth]{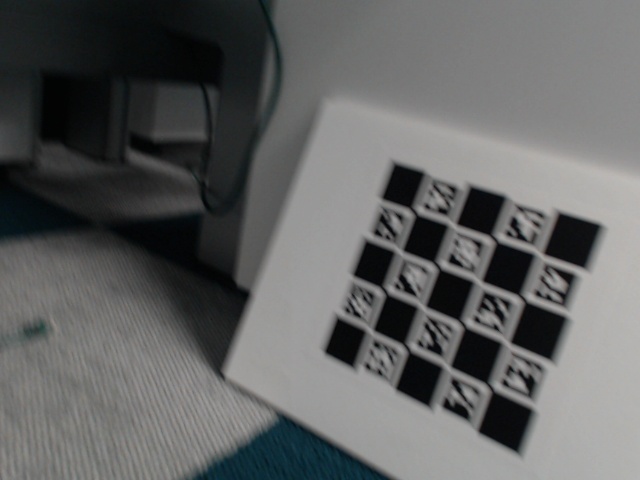}}
}
\caption{{\bf ChArUco Evaluation Set.} Examples of frames from the ChArUco evaluation set. From left to right, each frame focuses on lighting (10lux), shadow, motion blur.}
\label{fig:evaluation-set}
\end{figure}

For evaluation, we captured 26 videos of $1000$ frames at 30Hz from a Logitech\textsuperscript{\textregistered} webcam (see examples in Figure~\ref{fig:evaluation-set}). Each video in this set focuses on one of the following effects:
\begin{itemize}
    \vspace{-.05in}
	\item Lighting brightness (20 videos with 10 different lighting configurations)
    \vspace{-.05in}
	\item Shadow / spotlight (3 videos)
	\vspace{-.05in}
	\item Motion blur (3 videos)
\end{itemize}

\vspace{-.1in}

\begin{figure*}[tbp]
\vspace{-.25in}
\begin{center}
\includegraphics[width=0.99\textwidth]{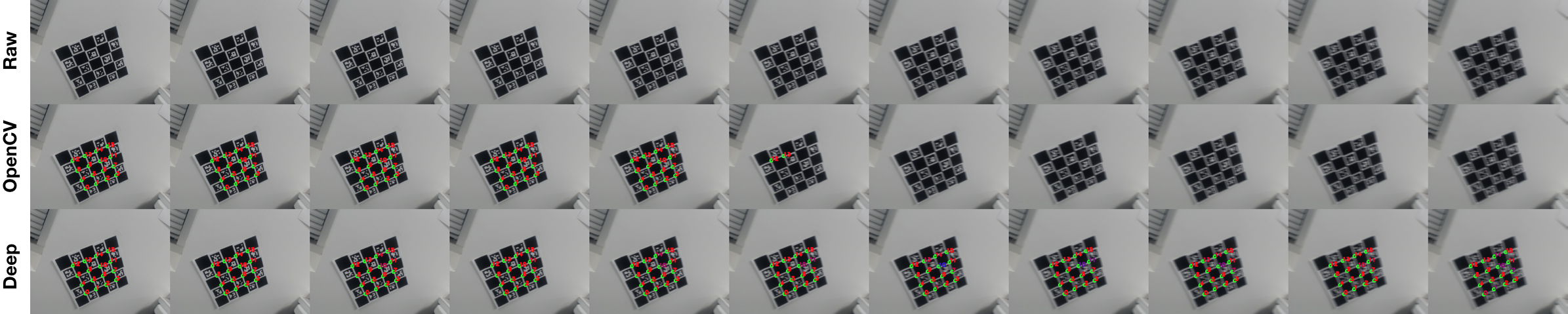}
\end{center}
\vspace{-.1in}
\caption{{\bf Synthetic Motion Blur Test Example.} Top row: input image applied with varying motion blur effect from kernel size 0 to 10; middle row: corners and ids detected by OpenCV detector, with detection accuracy [1.    1.    1.    1.    1.    0.125  0.  0.    0.    0.    0.    0.   ]; bottom row: corners and ids detected from the Deep ChArUco, with detection accuracy [1.     1.     1.     1.     1.     1.     1.     1.     1.  1.  1.  1.]}
\label{fig:motion_blur_example}
\end{figure*}

\section{Evaluation and Results} \label{sec:eval}
We compare our Deep ChArUco detector against a traditional OpenCV-based ChArUco marker detector in a frame-by-frame manner. We first evaluate both systems' ability to detect the $16$ ChArUco markers for a fixed set of images, under increasing blur and lighting changes (synthetic effects). Then, on real sequences, we estimate the pose of the ChArUco board based on the Perspective-n-Point algorithm and determine if the pose's reprojection error is below a threshold (typically $3$ pixels). Below, we outline the metrics used in our evaluation. 
\begin{itemize}
	\item Corner Detection Accuracy (accuracy of ChArUcoNet)
	\item ChArUco Pose Estimation Accuracy (combined accuracy of ChArUcoNet and RefineNet)
\end{itemize}

A corner is correctly detected when the location is within a $3$-pixel radius of the ground truth, and the point ID is identified correctly based on ChArUcoNet ID classifier. The \textbf{corner detection accuracy} is the ratio between the number of accurately detected corners and $16$, the total number of marker corners. The \textbf{average accuracy} is calculated as the mean of detection accuracy across $20$ images with different static poses. To quantitatively measure the \textbf{pose estimation accuracy} in each image frame, we use the mean reprojection error $\epsilon_{re}$ as defined below:
\begin{equation}
    \epsilon_{re} = \frac{\sum_{i=1}^{n}|\mathbf{PC}_i - c_i|}{n},
\end{equation}
where $\mathbf{P}$ is the camera projection matrix containing intrinsic parameters. $\mathbf{C}_i$ represents the 3D location of a detected corner computed from the ChArUco pose, $c_i$ denotes the 2d pixel location of the corresponding corner in the image. $n$ ($\leq 16$) is the total number of the detected ChArUco corners.

\begin{figure}[h]
\begin{center}
\includegraphics[width=0.9\linewidth]{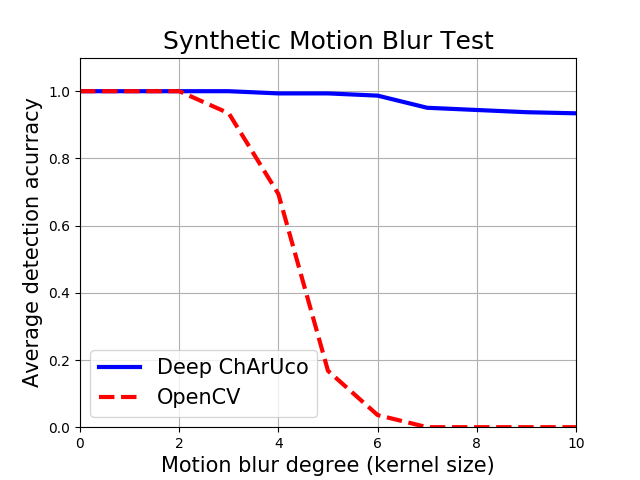}
\end{center}
\caption{{\bf Synthetic Motion Blur Test}. We compare Deep ChArUco with the OpenCV approach on $20$ random images from our test-set while increasing the amount of motion blur.}
\label{fig:motion_blur_acc}
\end{figure}

\subsection{Evaluation using synthetic effects}

In this section, we compare the overall accuracy of the Deep ChArUco detector and the OpenCV detector under synthetic effects, in which case, we vary the magnitude of the effect linearly. The first two experiments are aimed to evaluate the accuracy of ChArUcoNet output, without relying on RefineNet. 

In each of our $20$ synthetic test scenarios, we start with an image taken in an ideal environment - good lighting and random static pose (i.e., minimum motion blur), and gradually add synthetic motion blur and darkening.

\begin{figure*}[tbp]
\vspace{-.27in}
\begin{center}
\includegraphics[width=0.99\textwidth]{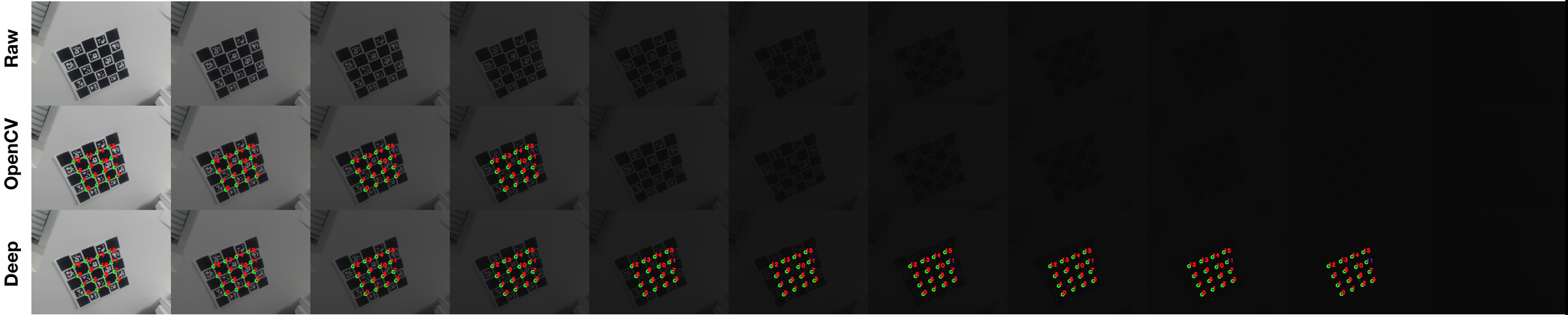}
\end{center}
\vspace{-.1in}
\caption{{\bf Synthetic Lighting Test Example.} Top row: input image applied with a brightness rescaling factor $0.6^k$ with k from 0 to 10; middle row: corners and ids detected by OpenCV detector with detection accuracy [1. 1. 1. 1. 0. 0. 0. 0. 0. 0. 0.]; bottom row: corners and ids detected from the Deep ChArUco with detection accuracy [1.     1.     1.     1.     1.     1.     1.     1.     1.  1.  0. ]}
\vspace{-.12in}
\label{fig:brightness_example}
\end{figure*}

\subsubsection{Synthetic Motion Blur Test}
In the motion blur test, a motion blur filter along the horizontal direction was applied to the original image with the varying kernel size to simulate the different degrees of motion blur. In Figure~\ref{fig:motion_blur_acc}, we plot average detection accuracy versus the degree of motion blur (i.e., the kernel size). It shows that Deep ChArUco is much more resilient to the motion blur effect compared to the OpenCV approach. Figure~\ref{fig:motion_blur_example} shows an example of increasing motion blur and the output of both detectors. Both the visual examples and resulting plot show that OpenCV methods start to completely fail ($0$\% detection accuracy) for kernel sizes of $6$ and larger, while Deep ChArUco only degrades a little bit in performance ($94$\% detection accuracy), even under extreme blur.

\subsubsection{Synthetic Lighting Test}
\begin{figure}[h]
\centering
\vspace{-.1in}
\includegraphics[width=0.9\linewidth]{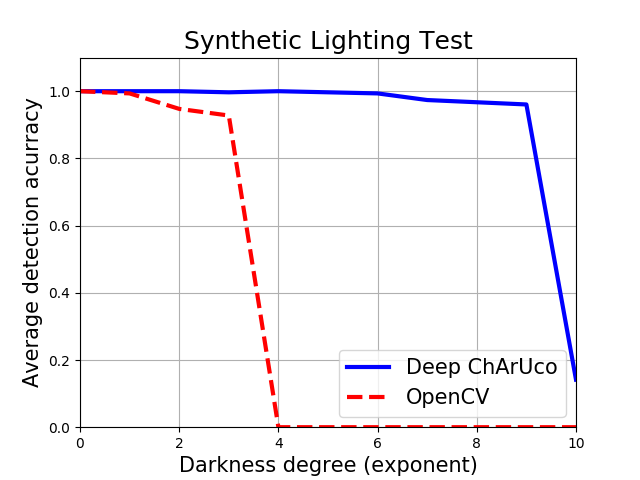}
\caption{{\bf Synthetic Lighting Test}.  We compare Deep ChArUco with the OpenCV approach on $20$ random images from our test-set while increasing the amount of darkness.}
\label{fig:brightness_test_acc}
\end{figure}

In the lighting test, we compare both detectors under different lighting conditions created synthetically. We multiply the original image with a rescaling factor of $0.6^k$ to simulate increasing darkness.
In Figure~\ref{fig:brightness_test_acc}, we plot average detection accuracy versus the darkness degree, $k$. Figure~\ref{fig:brightness_example} shows an example of increasing darkness and the output of both detectors. We note that Deep ChArUco is able to detect markers in many cases where the image is ``perceptually black'' (see last few columns of Figure~\ref{fig:brightness_example}). Deep ChArUco detects more than 50\% of the corners even when the brightness is rescaled by a factor of $0.6^9\sim.01$, while the OpenCV detector fails at the rescaling factor of $0.6^4\sim.13$. 

\vspace{-.13in}
\subsection{Evaluation on real sequences}
\begin{figure*}[tbp]
\centering
\vspace{-.33in}
\includegraphics[width=0.904\textwidth]{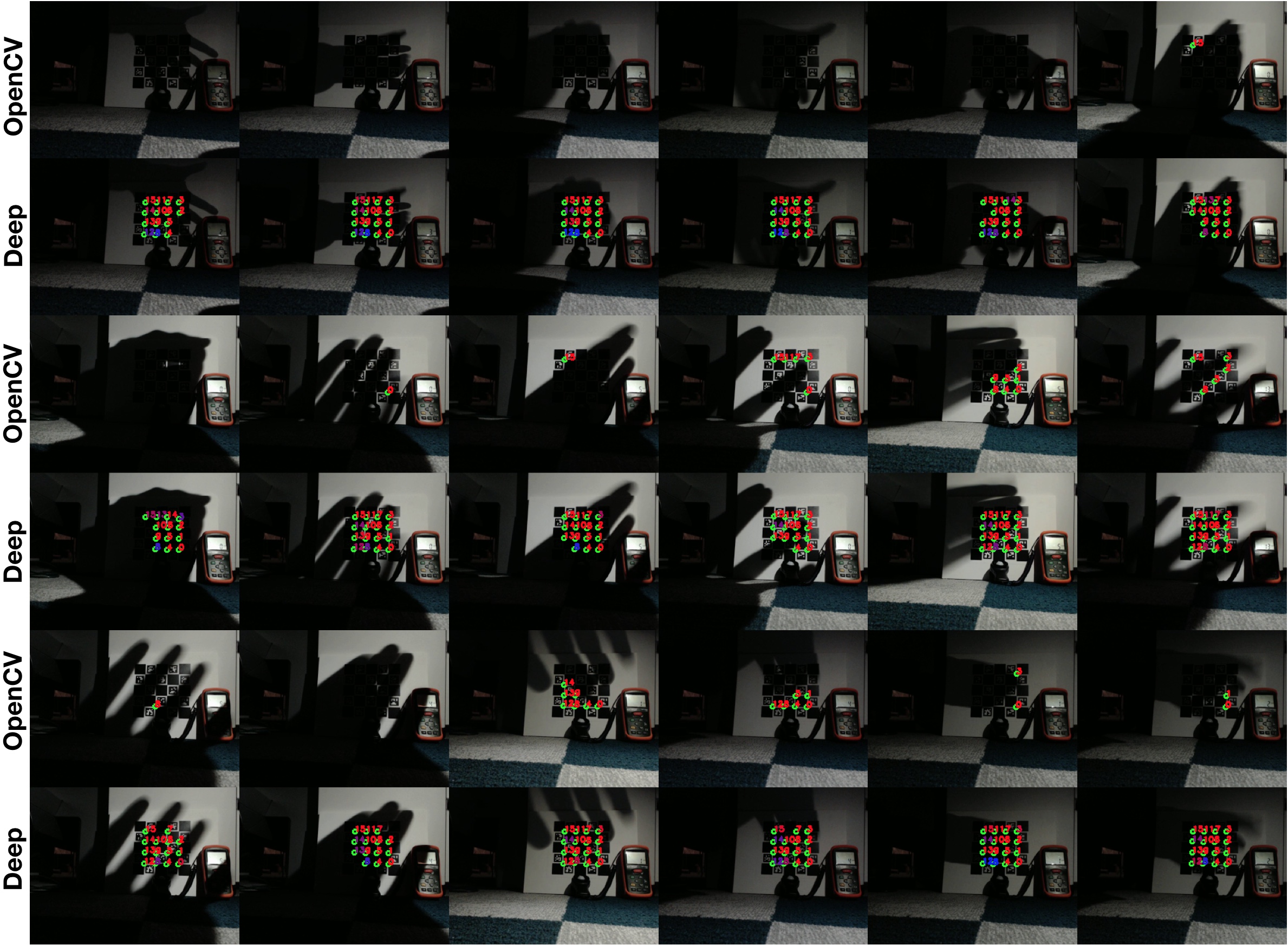} 
\includegraphics[width=0.907\textwidth]{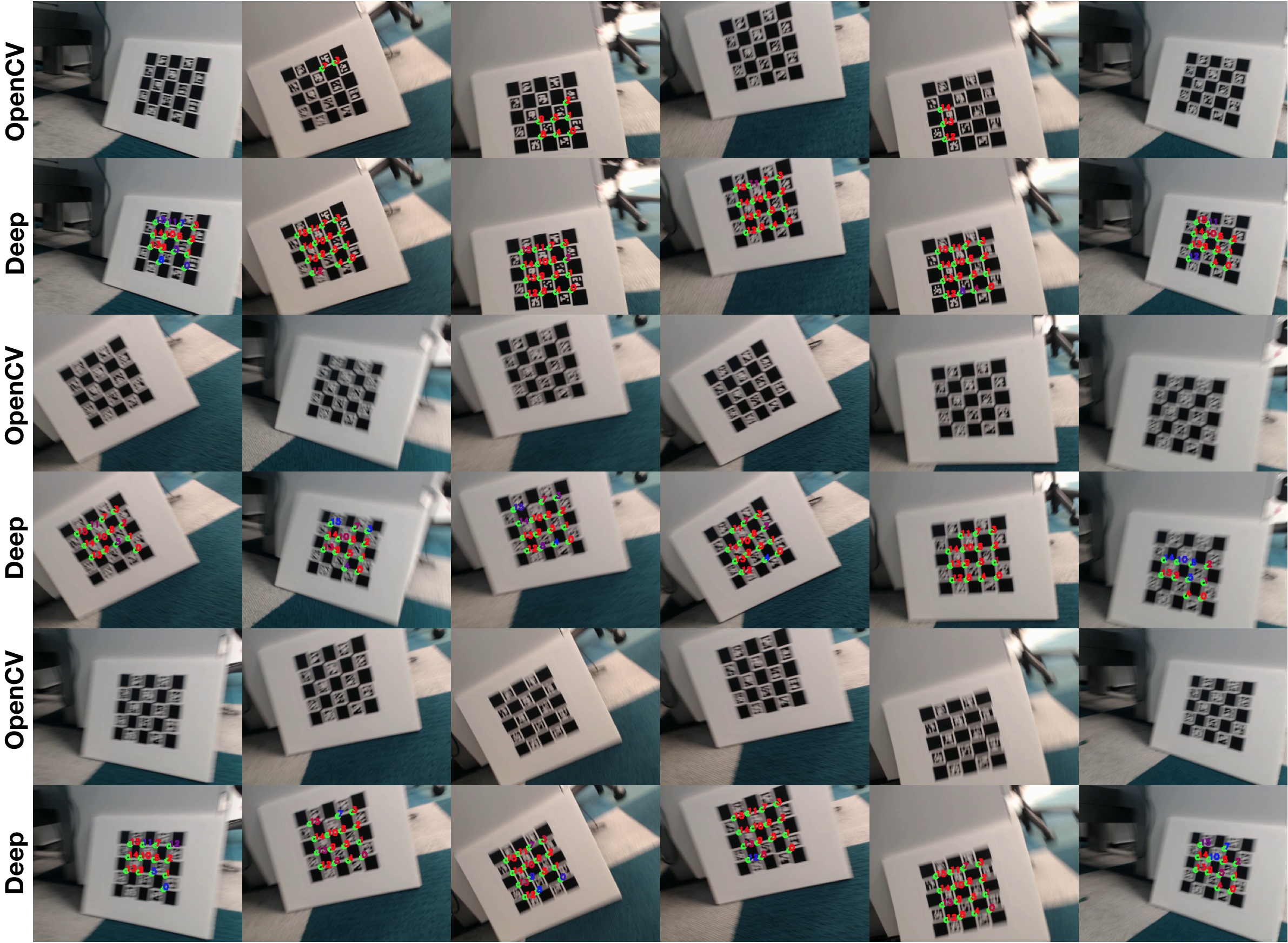} 
\caption{{\bf Deep ChArUco vs OpenCV Qualitative Examples.} Detector performance comparison under extreme lighting: shadows (top) and motion (bottom). Unlike OpenCV, Deep ChArUco appears unaffected by cast shadows.}
\label{fig:motion_light_combined}
\end{figure*}

First, we qualitatively show the accuracy of both detectors in real video clips captured in different scenarios as described in section \ref{sec:eval_data}, ``Evaluation Data.'' 
Figure~\ref{fig:motion_light_combined} shows the results of both detectors under extreme lighting and motion. Notice that the Deep ChArUco detector significantly outperforms the OpenCV detector under these extreme scenarios. Overall, our method detects more correct keypoints where a minimum number of $4$ correspondences is necessary for pose estimation.

\begin{figure}[h]
\begin{center}
\vspace{-.35in}
\includegraphics[width=0.99\linewidth]{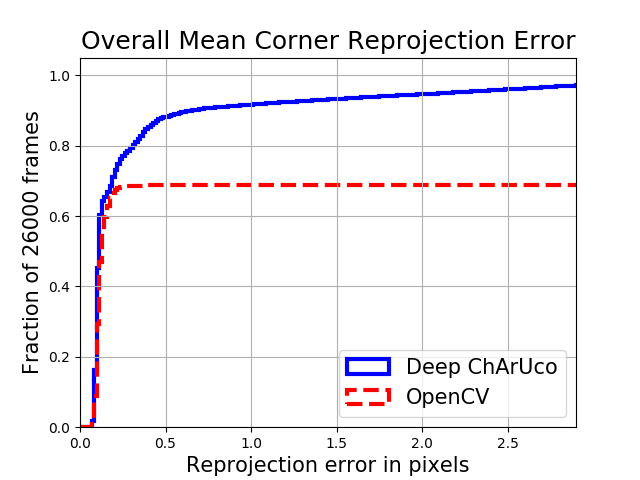}
\end{center}
\caption{{\bf Deep ChArUco vs OpenCV across entire evaluation dataset}. Pose accuracy vs. reprojection error $\epsilon_{re}$ threshold is computed across all $26,000$ frames in the $26$ videos of our evaluation set. Deep ChArUco exhibits higher pose estimation accuracy ($97.4\%$ vs. $68.8\%$ for OpenCV) under a $3$ pixel reprojection error threshold. }
\label{fig:hist}
\end{figure}

In our large experiment, we evaluate across all $26,000$ frames in the $26$-video dataset, without adding synthetic effects. We plot the fraction of correct poses vs. pose correctness threshold (as measured by reprojection error) in Figure~\ref{fig:hist}. Overall, we see that the Deep ChArUco system exhibits a higher detection rate ($97.4\%$ vs. $68.8\%$ under a $3$-pixel reprojection error threshold) and lower pose error compared to the traditional OpenCV detector. For each sequence in this experiment, Table~\ref{tab:individual_result} lists 
the ChArUco detection rate (where $\epsilon_{re} < 3.0$) and the mean $\epsilon_{re}$.

For sequences at $1$ and $0.3$ lux, OpenCV is unable to return a pose--they are too dark. For sequences with shadows, Deep ChArUco detects a good pose $100\%$ of the time, compared to $36\%$ for OpenCV. For videos with motion blur, Deep ChArUco works $78\%$ of the time, compared to $27\%$ for OpenCV. For a broad range of ``bright enough'' scenarios ranging from 3 lux to 700 lux, both Deep ChArUco and OpenCV successfully detect a pose $100$\% of the time, but Deep ChArUco has slightly lower reprojection error, $\epsilon_{re}$ on most sequences.\footnote{For per-video analysis on the $26$ videos in our evaluation dataset, please see the Appendix.}

\subsection{Deep ChArUco Timing Experiments}
\vspace{-.05in}
At this point, it is clear that Deep ChArUco works well under extreme lighting conditions, \emph{but is it fast enough for real-time applications?} We offer three options in network configuration based on the application scenarios with different requirements:
\begin{itemize}
\vspace{-.05in}
    \item \textbf{ChArUcoNet + RefineNet}: This is the recommended configuration for the best accuracy under difficult conditions like motion blur, low light, and strong imaging noise, but with longest post-processing time. 
    \vspace{-.05in}
    \item \textbf{ChArUcoNet + cornerSubPix}: For comparable accuracy in well-lit environment with less imaging noise, this configuration is recommended with moderate post-processing time.
    \vspace{-.05in}
    \item \textbf{ChArUcoNet + NoRefine}: This configuration is preferred when only the rough pose of the ChArUco pattern is required, especially in a very noisy environment where cornerSubPix will fail. The processing time is therefore the shortest as the image only passes through one CNN.
\end{itemize}

\begin{table}[h!]
\centering
\begin{tabular} {l|c}
\hline 
\textbf{Configurations} &  \textbf{Approx. fps (Hz)}\\
\hline\hline
ChArUcoNet + RefineNet & 66.5 \\
ChArUcoNet + cornerSubPix & 98.6 \\
ChArUcoNet + NoRefine & 100.7 \\
OpenCV detector + cornerSubPix & 99.4\\
OpenCV detector + NoRefine & 101.5 \\
\hline
\end{tabular}
\vspace{.05in}
\caption{{\bf Deep ChArUco Timing Experiments.} We present timing results for ChArUcoNet running on $320 \times 240$ images in three configurations: with RefineNet, with an OpenCV subpixel refinement step, and without refinement. Additionally, we also list the timing performance of OpenCV detector and refinement.}
\label{tab:processing_time}
\end{table}

We compare the average processing speed of $320 \times 240$ sized images using each of the above three configurations in Table~\ref{tab:processing_time}. The reported framerate is an average across the evaluation videos described in Section~\ref{sec:eval_data}. Experiments are performed using an NVIDIA\textsuperscript{\textregistered} GeForce GTX 1080 GPU. Since ChArUcoNet is fully convolutional, it is possible to apply the network to different image resolutions, depending on computational or memory requirements. To achieve the best performance with larger resolution images, we can pass a low-resolution image through ChArUcoNet to roughly localize the pattern and then perform subpixel localization via RefineNet in the original high-resolution image. 

\begin{table}[ht!]
\vspace{-.35in}
\centering
\scalebox{0.95}{
\begin{tabular} {|l|c|c|c|c|}
\hline 
\textbf{Video} &  \textbf{deep acc} & \textbf{cv acc} & \textbf{deep $\epsilon_{re}$}  & \textbf{cv $\epsilon_{re}$}\\
\hline
0.3lux & \bf{100} & 0 & \bf{0.427} (0.858) & nan \\
0.3lux & \bf{100} & 0 & \bf{0.388} (0.843) & nan \\
1lux & \bf{100} & 0 & \bf{0.191} (0.893) & nan \\
1lux & \bf{100} & 0 & \bf{0.195} (0.913) & nan \\
\hdashline
3lux & 100 & 100 & \textbf{0.098} (0.674) & 0.168 \\
3lux & 100 & 100 & \textbf{0.097} (0.684) & 0.164 \\
5lux & 100 & 100 & \textbf{0.087} (0.723) & 0.137 \\
5lux & 100 & 100 & \textbf{0.091} (0.722) & 0.132 \\
10lux & 100 & 100 & \textbf{0.098} (0.721) & 0.106 \\
10lux & 100 & 100 & \textbf{0.097} (0.738) & 0.105 \\
30lux & 100 & 100 & 0.100 (0.860) & \textbf{0.092} \\
30lux & 100 & 100 & 0.100 (0.817) & \textbf{0.088}\\
50lux & 100 & 100 & 0.103 (0.736) & \textbf{0.101} \\
50lux & 100 & 100 & 0.102 (0.757) & \textbf{0.099} \\
100lux & 100 & 100 & 0.121 (0.801) & \textbf{0.107} \\
100lux & 100 & 100 & \textbf{0.100} (0.775) & 0.118 \\
400lux & 100 & 100 & \textbf{0.086} (0.775) & 0.093 \\
400lux & 100 & 100 & \textbf{0.085} (0.750) & 0.093 \\
700lux & 100 & 100 & \textbf{0.102} (0.602) & 0.116 \\
700lux & 100 & 100 & \textbf{0.107} (0.610) & 0.120 \\
\hdashline
shadow 1 & \bf{100} & 42.0 & 0.254 (0.612) & 0.122 \\
shadow 2 & \bf{100} & 30.1 & 0.284 (0.618) & 0.130 \\
shadow 3 & \bf{100} & 36.9 & 0.285 (0.612) & 0.141\\
\hdashline
motion 1 & \bf{74.1} & 16.3 & 1.591 (0.786) & 0.154 \\
motion 2 & \bf{78.8} & 32.1 & 1.347 (0.788) & 0.160\\
motion 3 & \bf{80.3} & 31.1 & 1.347 (0.795) & 0.147\\
\hline
\end{tabular}}
\vspace{.05in}
\caption{{\bf Deep ChArUco vs OpenCV Individual Video Summary.} We report the pose detection accuracy (percentage of frames with reprojection error less than $3$ pixels) as well as the mean reprojection error, $\epsilon_{re}$, for each of our $26$ testing sequences. Notice that OpenCV is unable to return a marker pose for images at 1 lux or darker (indicated by nan). The deep reprojection error column also lists the error without RefineNet in parenthesis. RefineNet reduces the reprojection error in all cases except the motion blur scenario, because in those cases the ``true corner'' is outside of the central $8 \times 8$ refinement region.}
\label{tab:individual_result}
\vspace{-.1in}
\end{table}


\vspace{-.05in}
\section{Conclusion}
\vspace{-.05in}
\label{sec:conclusions}

Our paper demonstrates that deep convolutional neural networks can dramatically improve the detection rate for ChArUco markers in low-light, high-motion scenarios where the traditional ChArUco marker detection tools inside OpenCV often fail.  We have shown that our Deep ChArUco system, a combination of ChArUcoNet and RefineNet, can match or surpass the pose estimation accuracy of the OpenCV detector. Our synthetic and real-data experiments show a performance gap favoring our approach and demonstrate the effectiveness of our neural network architecture design and the dataset creation methodology. The key ingredients to our method are the following: ChArUcoNet, a CNN for pattern-specific keypoint detection, RefineNet, a subpixel localization network, a custom ChArUco pattern-specific dataset, comprising extreme data augmentation and proper selection of visually similar patterns as negatives. Our system is ready for real-time applications requiring marker-based pose estimation.

Furthermore, we used a specific ChArUco marker as an example in this work. By replacing the ChArUco marker with another pattern and collecting a new dataset (with manual labeling if the automatic labeling is too hard to achieve), the same training procedure could be repeated to produce numerous pattern-specific networks. Future work will focus on multi-pattern detection, end-to-end learning, and pose estimation of non-planar markers.

\clearpage
\newpage

{\normalsize 
\bibliographystyle{ieee_fullname} 
\bibliography{main}

\begin{thebibliography}{10}\itemsep=-1pt

\bibitem{bradski2000opencv}
Gary Bradski and Adrian Kaehler.
\newblock Opencv.
\newblock {\em Dr. Dobb’s journal of software tools}, 3, 2000.

\bibitem{cao2017realtime}
Zhe Cao, Tomas Simon, Shih-En Wei, and Yaser Sheikh.
\newblock Realtime multi-person 2d pose estimation using part affinity fields.
\newblock In {\em CVPR}, 2017.

\bibitem{fiducialsfm}
Joseph DeGol, Timothy Bretl, and Derek Hoiem.
\newblock Improved structure from motion using fiducial marker matching.
\newblock In {\em ECCV}, 2018.

\bibitem{detone2018}
Daniel DeTone, Tomasz Malisiewicz, and Andrew Rabinovich.
\newblock Superpoint: Self-supervised interest point detection and description.
\newblock In {\em CVPR Deep Learning for Visual SLAM Workshop}, 2018.

\bibitem{artag}
Mark Fiala.
\newblock Artag, a fiducial marker system using digital techniques.
\newblock In {\em CVPR}, 2005.

\bibitem{garrido2014}
Sergio Garrido-Jurado, Rafael Mu{\~n}oz-Salinas, Francisco~Jos{\'e}
  Madrid-Cuevas, and Manuel~Jes{\'u}s Mar{\'\i}n-Jim{\'e}nez.
\newblock Automatic generation and detection of highly reliable fiducial
  markers under occlusion.
\newblock {\em Pattern Recognition}, 47(6):2280--2292, 2014.

\bibitem{hz}
Richard Hartley and Andrew Zisserman.
\newblock {\em Multiple view geometry in computer vision}.
\newblock Cambridge university press, 2003.

\bibitem{maskrcnn}
Kaiming He, Georgia Gkioxari, Piotr Doll{\'a}r, and Ross Girshick.
\newblock Mask r-cnn.
\newblock In {\em ICCV}, 2017.

\bibitem{fiducialslam2}
Hyon Lim and Young~Sam Lee.
\newblock Real-time single camera slam using fiducial markers.
\newblock In {\em ICCAS-SICE}, 2009.

\bibitem{Wei2016}
Wei Liu, Dragomir Anguelov, Dumitru Erhan, Christian Szegedy, Scott Reed,
  Cheng-Yang Fu, and Alexander~C Berg.
\newblock Ssd: Single shot multibox detector.
\newblock In {\em ECCV}, 2016.

\bibitem{sift}
David~G Lowe.
\newblock Distinctive image features from scale-invariant keypoints.
\newblock {\em International journal of computer vision}, 60(2):91--110, 2004.

\bibitem{aruco}
Rafael Munoz-Salinas.
\newblock Aruco: a minimal library for augmented reality applications based on
  opencv.
\newblock {\em Universidad de C{\'o}rdoba}, 2012.

\bibitem{hourglass}
Alejandro Newell, Kaiyu Yang, and Jia Deng.
\newblock Stacked hourglass networks for human pose estimation.
\newblock In {\em ECCV}, 2016.

\bibitem{apriltag}
Edwin Olson.
\newblock Apriltag: A robust and flexible visual fiducial system.
\newblock In {\em ICRA}, 2011.

\bibitem{yolo}
Joseph Redmon, Santosh Divvala, Ross Girshick, and Ali Farhadi.
\newblock You only look once: Unified, real-time object detection.
\newblock In {\em CVPR}, 2016.

\bibitem{Ren2015}
Shaoqing Ren, Kaiming He, Ross Girshick, and Jian Sun.
\newblock Faster r-cnn: Towards real-time object detection with region proposal
  networks.
\newblock In {\em NIPS}, 2015.

\bibitem{orb}
Ethan Rublee, Vincent Rabaud, Kurt Konolige, and Gary Bradski.
\newblock Orb: An efficient alternative to sift or surf.
\newblock In {\em ICCV}, 2011.

\bibitem{simon2017hand}
Tomas Simon, Hanbyul Joo, Iain~A Matthews, and Yaser Sheikh.
\newblock Hand keypoint detection in single images using multiview
  bootstrapping.
\newblock In {\em CVPR}, 2017.

\bibitem{posecnn}
Yu Xiang, Tanner Schmidt, Venkatraman Narayanan, and Dieter Fox.
\newblock Posecnn: A convolutional neural network for 6d object pose estimation
  in cluttered scenes.
\newblock {\em Robotics: Science and Systems (RSS)}, 2018.

\bibitem{Moo2016}
Kwang~Moo Yi, Eduard Trulls, Vincent Lepetit, and Pascal Fua.
\newblock Lift: Learned invariant feature transform.
\newblock In {\em ECCV}, 2016.

\end{thebibliography}
}

\clearpage

\appendix

\vspace{.1in}

\begin{figure*}[th!]
\vspace{-.6in}
\noindent{\huge\bfseries Appendix\par}
    \centering
    \begin{tabular}{ccc}
    \includegraphics[width=0.33\textwidth]{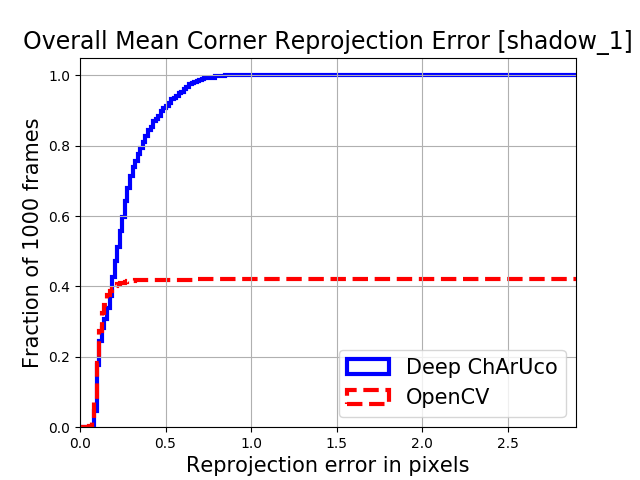} &
    \includegraphics[width=0.33\textwidth]{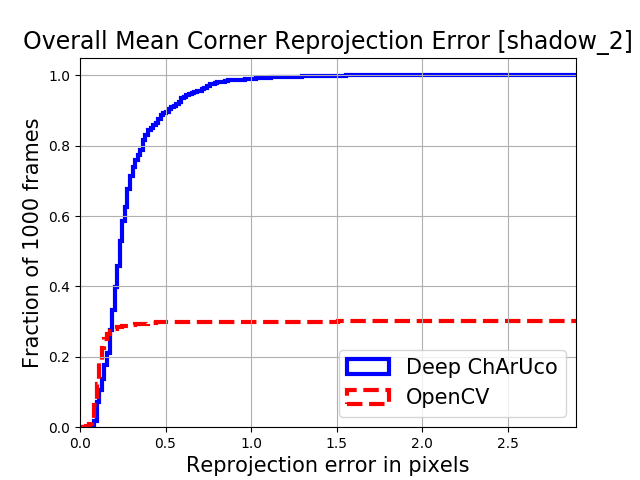} &
    \includegraphics[width=0.33\textwidth]{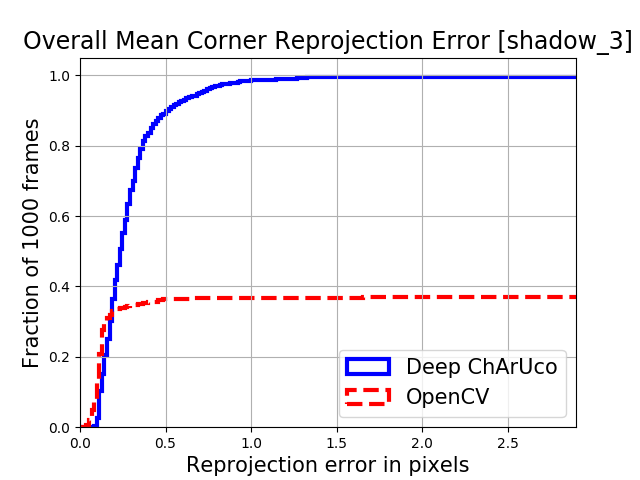} \\
    \end{tabular}
    \caption{{\bf Shadow Sequences}. We report the pose accuracy vs. reprojection error threshold on the following sequences: \texttt{shadow\_1}, \texttt{shadow\_2}, and \texttt{shadow\_3}. The results on shadow sequences indicate that Deep ChArUco is very robust to nuisance factors such as cast shadows. See top of Figure~\ref{fig:motion_light_combined} for examples of difficult shadows--Deep ChArUco is relatively unaffected by such shadows while OpenCV rarely detects point IDs behind a shadow.}
    \label{fig:shadow_all}
\end{figure*}   

\begin{figure*}[th!]
    \centering
    \begin{tabular}{ccc}
    \includegraphics[width=0.33\textwidth]{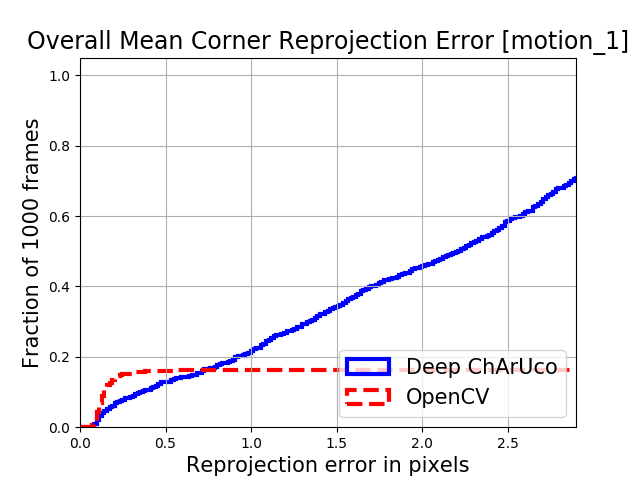} &
    \includegraphics[width=0.33\textwidth]{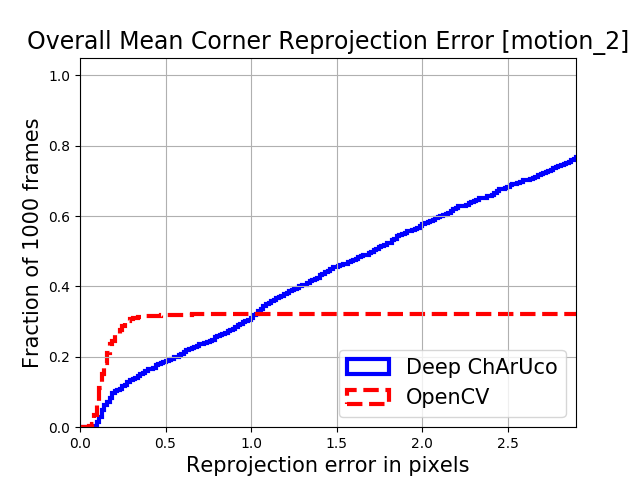} &
    \includegraphics[width=0.33\textwidth]{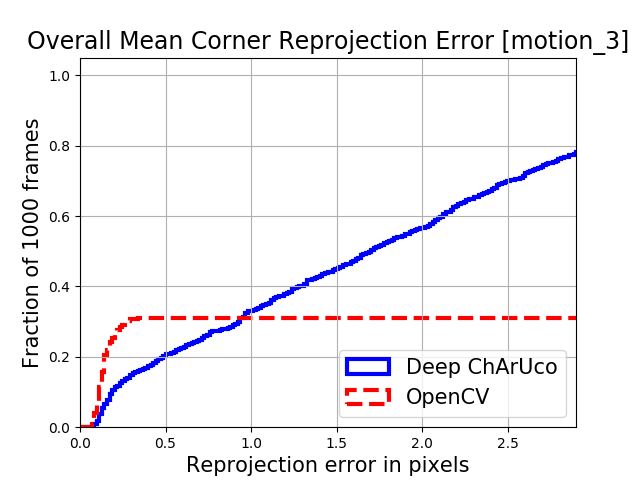} \\
    \end{tabular}
    \caption{{\bf Motion blur Sequences}. We report the pose accuracy vs. reprojection error threshold on the following sequences: \texttt{motion\_1}, \texttt{motion\_2}, and \texttt{motion3\_3} (see bottom of Figure~\ref{fig:motion_light_combined}). For motion blur sequences, we see that the traditional method is slightly better when a pose threshold of $1$ pixel or less is chosen. This suggests that Deep ChArUco could benefit from training with examples of real (non-synthetic) blur.}
    \label{fig:motion_all}
\end{figure*}  

\begin{figure*}[h]
    \centering
    \begin{tabular}{cc}
      \includegraphics[width=0.33\textwidth]{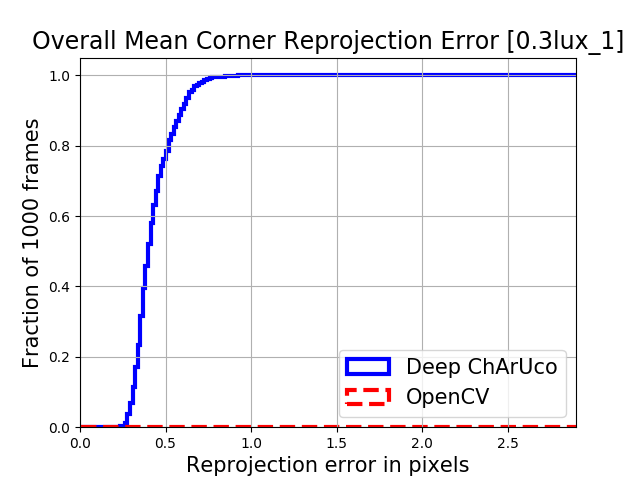}  & \includegraphics[width=0.33\textwidth]{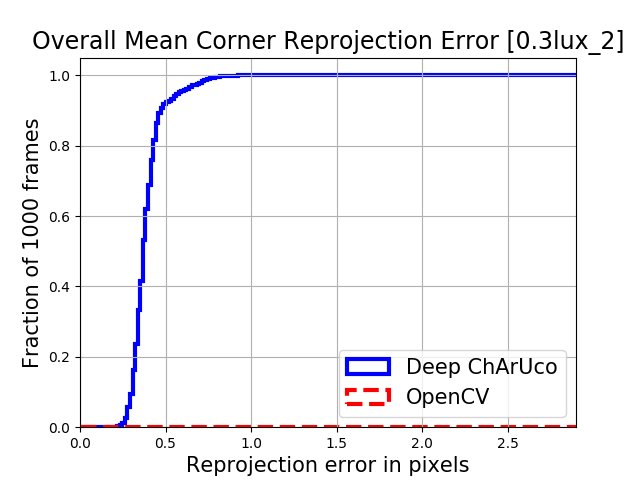} \\
      \includegraphics[width=0.33\textwidth]{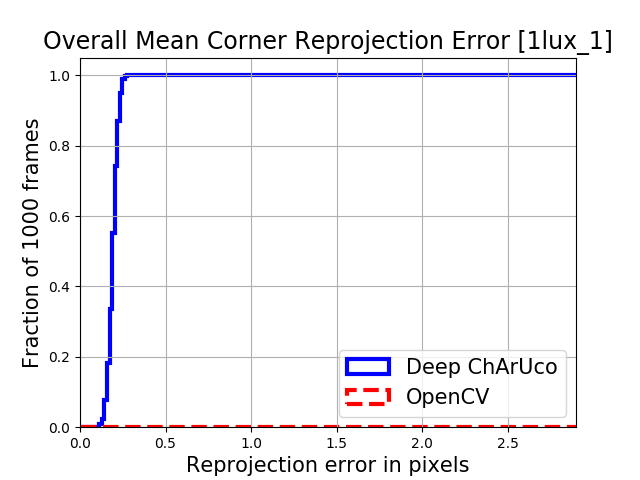}   & 
      \includegraphics[width=0.33\textwidth]{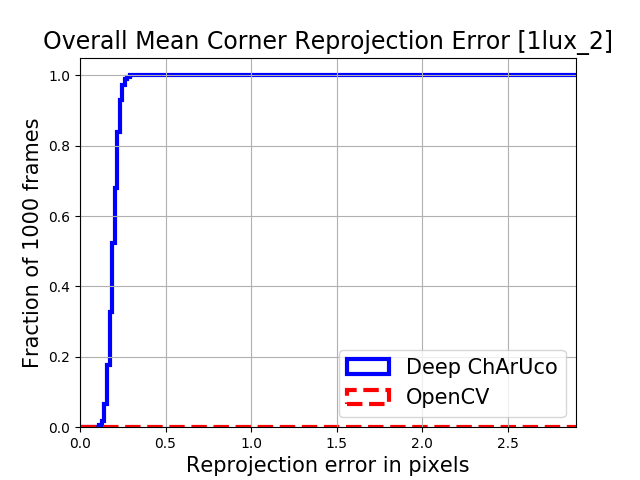}
    \end{tabular}
    \caption{{\bf Extreme Low Light Sequences.} We report the pose accuracy vs. reprojection error threshold on the sequences at $1$ lux and below. OpenCV completely fails. }
    \label{fig:lowlight_all}
\end{figure*}

\begin{figure*}[h]
\vspace{-.8in}
    \centering
    \begin{tabular}{ccc}
    \includegraphics[width=0.3\textwidth]{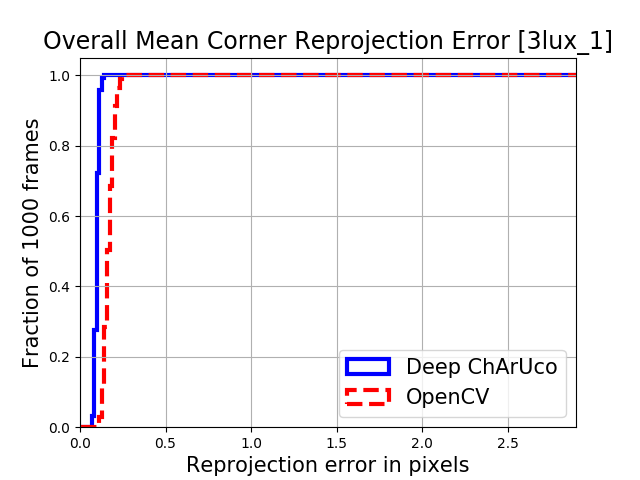} &
    \includegraphics[width=0.3\textwidth]{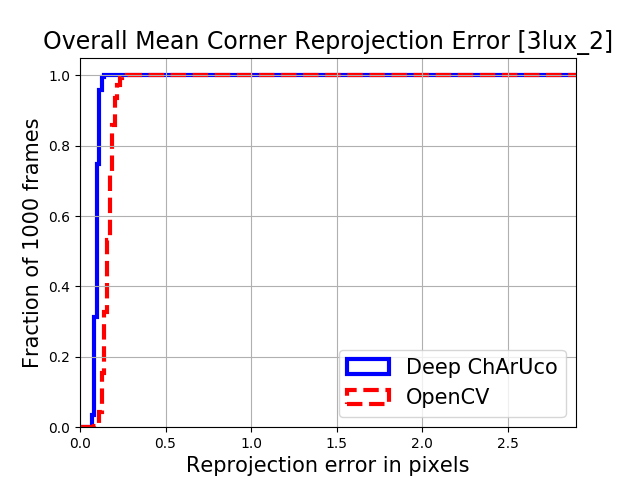} &
    \includegraphics[width=0.3\textwidth]{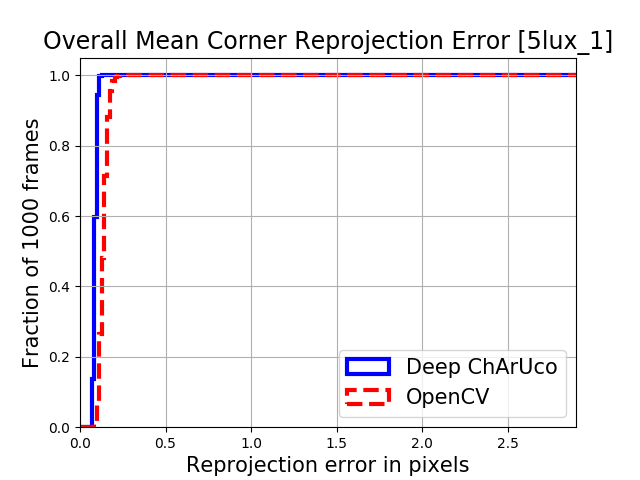} \\
    \includegraphics[width=0.3\textwidth]{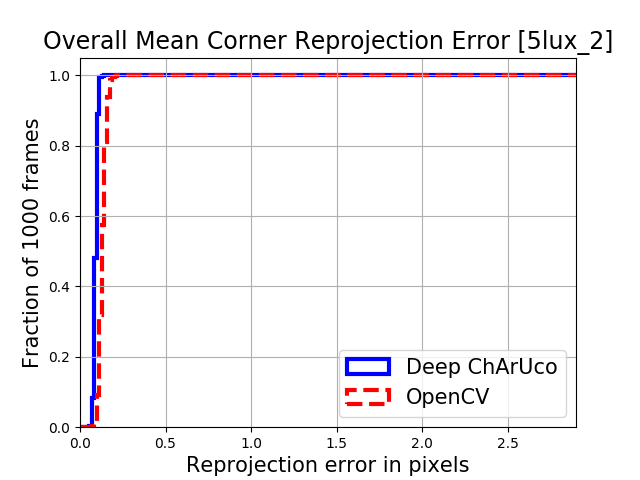} &
    \includegraphics[width=0.3\textwidth]{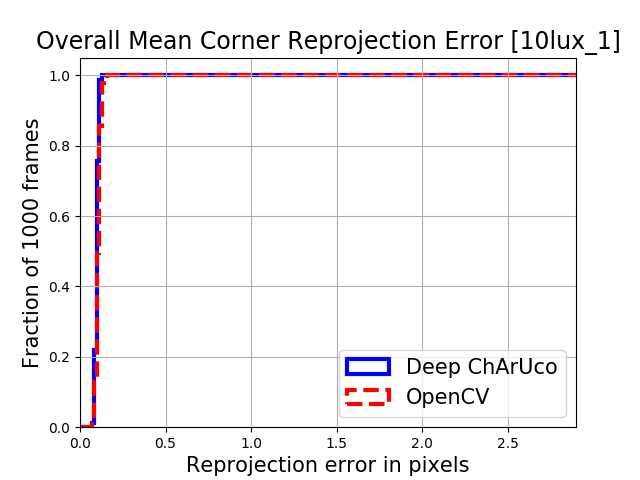} & 
    \includegraphics[width=0.3\textwidth]{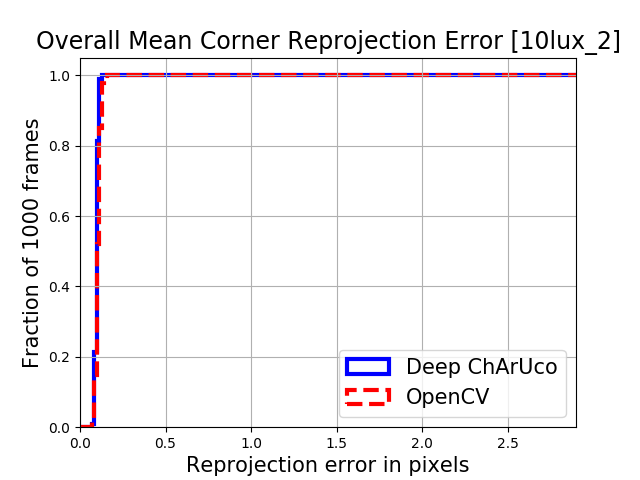} \\
    \includegraphics[width=0.3\textwidth]{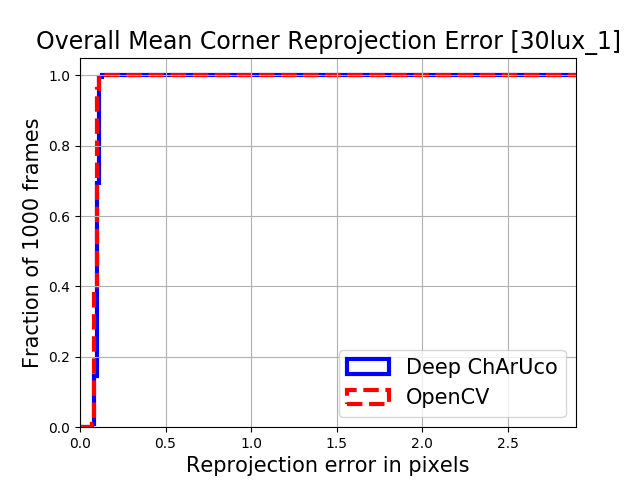} &
    \includegraphics[width=0.3\textwidth]{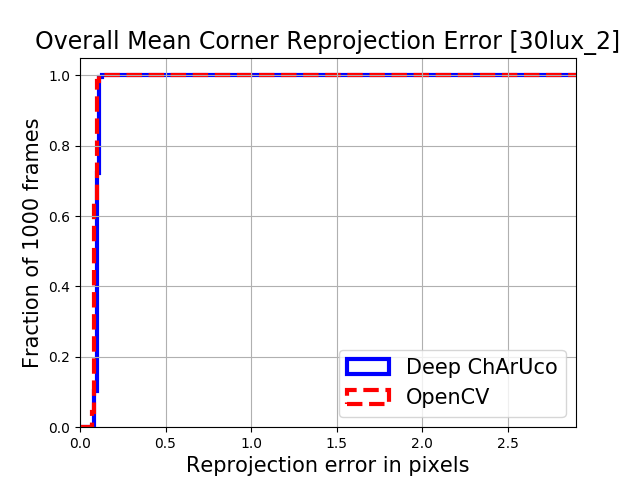} &
    \includegraphics[width=0.3\textwidth]{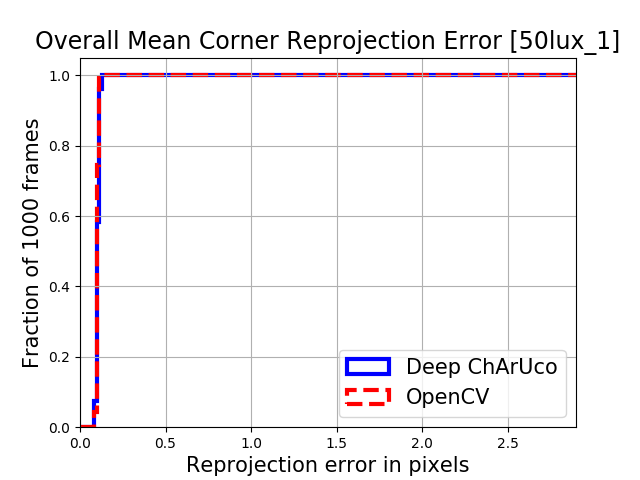} \\
    \includegraphics[width=0.3\textwidth]{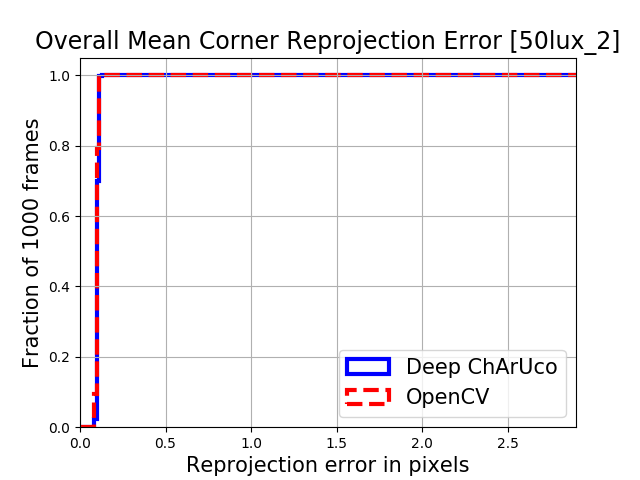} &
    \includegraphics[width=0.3\textwidth]{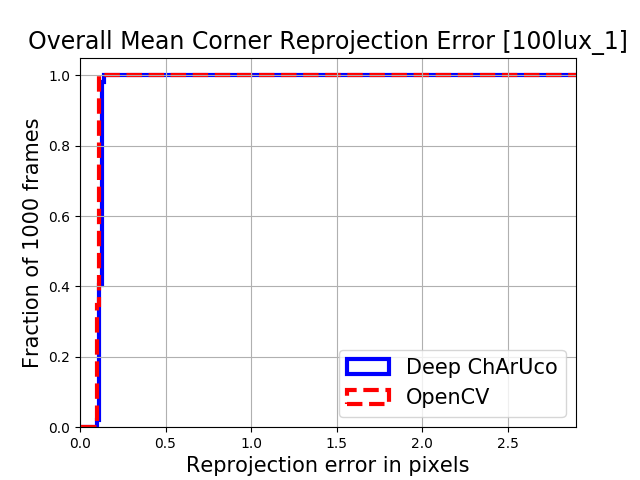} &
    \includegraphics[width=0.3\textwidth]{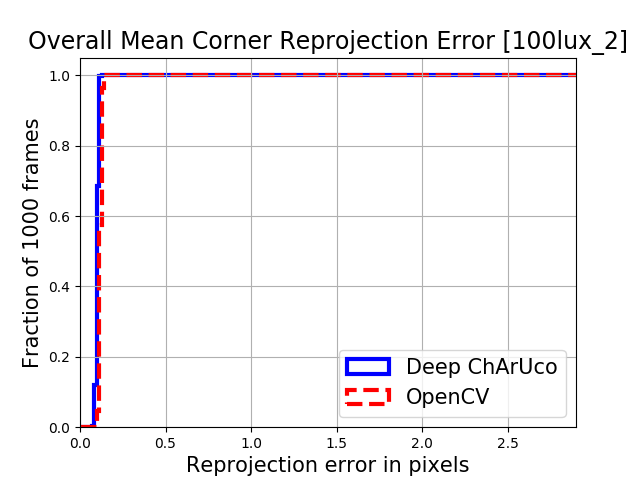} \\
    \includegraphics[width=0.3\textwidth]{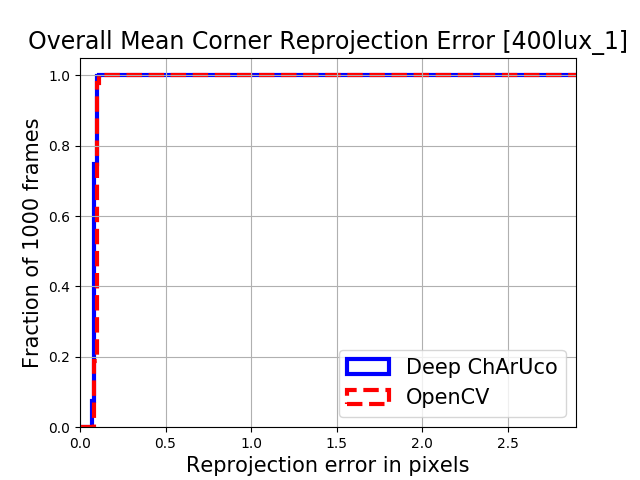} &
    \includegraphics[width=0.3\textwidth]{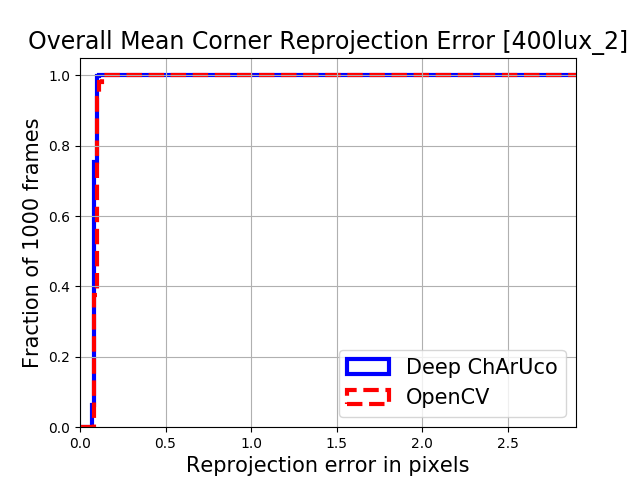} &
    \includegraphics[width=0.3\textwidth]{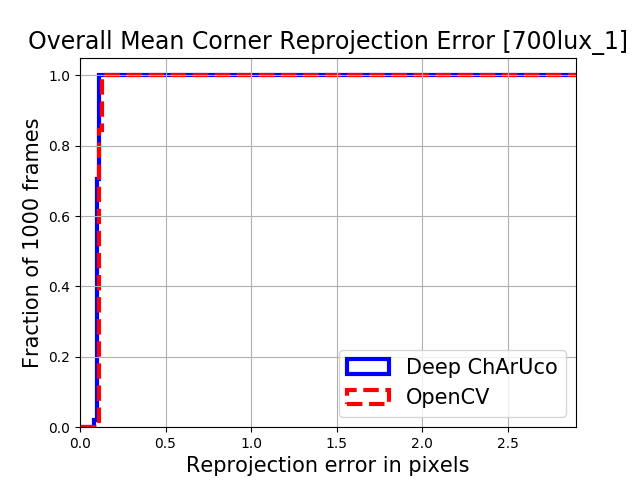} \\
    \includegraphics[width=0.3\textwidth]{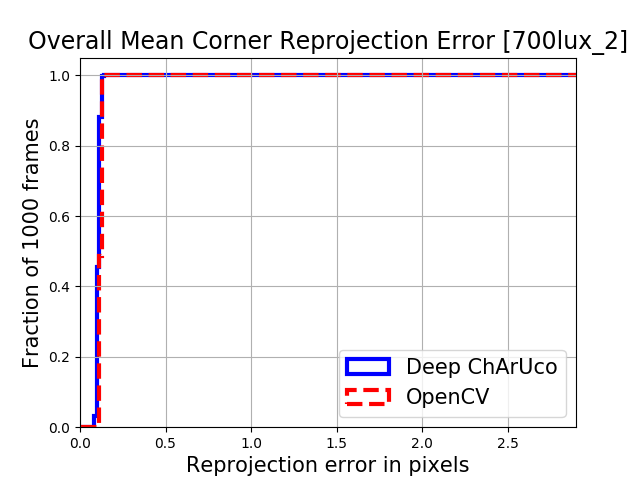} 
    \end{tabular}
    \caption{{\bf Normal Light Sequences.} We report the pose accuracy vs. reprojection error threshold on the sequences between $3$ and $700$ lux. From $3$ to $5$ lux, Deep ChArUco shows a visible improvement, while for a higher lux, both methods perform similarly.}
    \label{fig:normal_all}
\end{figure*}   

\end{document}